\documentclass[sn-mathphys-num]{sn-jnl}

\usepackage{graphicx}%
\usepackage{multirow}%
\usepackage{amsmath,amssymb,amsfonts}%
\usepackage{amsthm}%
\usepackage{mathrsfs}%
\usepackage[title]{appendix}%
\usepackage{xcolor}%
\usepackage{textcomp}%
\usepackage{manyfoot}%
\usepackage{booktabs}%
\usepackage{algorithm}%
\usepackage{algorithmicx}%
\usepackage{algpseudocode}%
\usepackage{listings}%

\usepackage{url}
\usepackage{hyperref}

\theoremstyle{thmstyleone}%
\theoremstyle{thmstyletwo}%

\theoremstyle{thmstylethree}%

\begin{document}

\title[]{Image Retrieval with Intra-Sweep Representation Learning for Neck Ultrasound Scanning Guidance}


\author*[1]{\fnm{Wanwen} \sur{Chen}}\email{wanwenc@ece.ubc.ca}
\author[1,2]{\fnm{Adam} \sur{Schmidt}}
\author[3]{\fnm{Eitan} \sur{Prisman}}

\author[1,4]{\fnm{Septimiu E.} \sur{Salcudean}}

\affil*[1]{\orgdiv{Department of Electrical and Computer Engineering}, \orgname{The University of British Columbia}, \orgaddress{\city{Vancouver}, \state{BC}, \country{Canada}}}

\affil[2]{\orgname{Intuitive Surgical Inc.}, \orgaddress{\city{Vancouver}, \state{BC}, \country{Canada}}}

\affil[3]{\orgdiv{Division of Otolaryngology, Department of Surgery}, \orgname{The University of British Columbia}, \orgaddress{\city{Vancouver}, \state{BC}, \country{Canada}}}

\affil[4]{\orgdiv{School of Biomedical Engineering}, \orgname{The University of British Columbia}, \orgaddress{\city{Vancouver}, \state{BC}, \country{Canada}}}



\abstract{\textbf{Purpose:} Intraoperative ultrasound (US) can enhance real-time visualization in transoral robotic surgery. The surgeon creates a mental map with a pre-operative scan. Then, a surgical assistant performs freehand US scanning during the surgery while the surgeon operates at the remote surgical console. Communicating the target scanning plane in the surgeon's mental map is difficult. 
Automatic image retrieval can help match intraoperative images to preoperative scans, guiding the assistant to adjust the US probe toward the target plane.

\textbf{Methods:} We propose a self-supervised contrastive learning approach to match intraoperative US views to a preoperative image database. We introduce a novel contrastive learning strategy that leverages intra-sweep similarity and US probe location to improve feature encoding. Additionally, our model incorporates a flexible threshold to reject unsatisfactory matches. 

\textbf{Results:} 
Our method achieves 92.30\% retrieval accuracy on simulated data and outperforms state-of-the-art temporal-based contrastive learning approaches. Our ablation study demonstrates that using probe location in the optimization goal improves image representation, suggesting that semantic information can be extracted from probe location. We also present our approach on real patient data to show the feasibility of the proposed US probe localization system despite tissue deformation from tongue retraction.

\textbf{Conclusion:}  Our contrastive learning method, which utilizes intra-sweep similarity and US probe location, enhances US image representation learning.  We also demonstrate the feasibility of using our image retrieval method to provide neck US localization on real patient US after tongue retraction. 
}




\maketitle

\section{Introduction}\label{sec1}
With the advances in medical robot technology, trans-oral robotic surgery (TORS) using the da Vinci surgical robot has become a recommended treatment for early-stage oropharyngeal cancers under the National Comprehensive Cancer Network guidelines~\cite{adelstein2017nccn}. 
TORS has resulted in similar oncological cure rates while providing improved functional outcomes compared to radiotherapy~\cite{quan2021gastrostomy,lechner2022hpv}. 
During TORS, surgeons need to accurately remove the cancer. This requires balancing maintaining a negative margin while avoiding critical structures and preserving as much healthy tissue as possible. 
TORS is a challenging procedure because of the proximity of cancerous tumors to sensitive anatomy, including major vasculature.
Surgeons can only refer to pre-operative MRI and/or CT, but pre-operative imaging can not show the tissue deformation caused by tongue retraction and surgery tools.
Consequently, real-time image guidance is needed to display the patient’s internal anatomy to enhance the surgeon's visualization of both the tumor margin and critical structures. 
This can help the surgeon optimize the resection margin and reduce the risk of postoperative hemorrhage. 
Such image guidance has been proposed using X-ray C-arms and cone-beam CT~\cite{liu2015augmented,kahng2019improving} or US~\cite{chen2023towards}.
Oropharyngeal ultrasound (US)~\cite{green2020integrated,chang2021real} is a new imaging technique recently introduced to guide TORS. 
Compared to CT, it has better portability, lower cost, and does not introduce harmful radiation. 

The clinical workflow for US-guided TORS requires collaboration between the surgeon and a surgical assistant. 
After the anesthetization, the surgeon and the surgical assistant perform a thorough pre-operative US scan together to identify important views that visualize tumors and critical structures. 
During the surgery, the surgeon operates the surgical robot remotely and can only see the 2D US in the console. 
The surgical assistant sits by the patient to perform real-time freehand US scanning. 
Thus, the surgeon needs to communicate the target US plane verbally. 
Moore {\em et al.}~\cite{moore2024enabling} proposed using the robotic arm on the da Vinci to let the surgeon control the US probe directly.
However, obtaining regulatory approval for such an off-label use of the da Vinci system could be difficult. 
Therefore, at this point, freehand scanning by a surgical assistant is still preferable for safety concerns. 
However, US has a limited field of view and US scanning has a steep learning curve. 
The surgical assistants are usually surgical residents or fellows instead of radiologists, thus a US scanning guidance system can further ease their mental load.

Previous work in US scanning guidance usually predicts the current US probe location or ideal US probe movement to a {\em target plane}. 
Grimwood {\em et al.}~\cite{grimwood2020assisted} provide US operator assistance in the form of motion directions instead of precise target probe movement.
Droste {\em et al.}~\cite{droste2020automatic} developed a model to predict target rotation with US and inertial measurement unit (IMU) signals as extra inputs for obstetric scanning.
Zhao {\em et al.}~\cite{zhao2022uspoint}  used a keypoint detection and matching model to predict probe motion for fine adjustment in standard plane search.
Most recently, Men {\em et al.}~\cite{men2024pose} developed a model that can predict pose and scanning guidance without external sensors.
However, previous work was mostly applied to echocardiography and fetal US, in which the {\em standard planes} are well defined.
Head and neck US scans do not have a similar definition of the standard plane; the surgeons need to examine different views during the surgery for a more comprehensive understanding of the anatomy near the tumor.

In this paper, we propose using US image retrieval for US probe localization and guidance.
Image retrieval is the task of finding the most similar image in an image database.
In our work, the image database comprises the reference frames and their probe locations in the pre-operative scans.
Compared with estimating and predicting probe location, image retrieval models have the potential to generalize across different patients since they use patient-specific reference images along with each patient's database. 
However, image retrieval is not widely applied in US scanning guidance.
In the limited prior work, Zhao {\em et al.}~\cite{zhao2021visual} have applied image (landmark) retrieval in fetal US scanning.
However, their data were generated through simulation, without considering the information embedded in the actual US scan videos.
Yeung et al.~\cite{yeung2021learning} resliced 3D US into 2D given a simulated probe location and trained the model to predict the probe location using supervised learning.
However, the real 2D US and the resliced 2D US have different appearances, which can impact the model performance, so in their follow-up work~\cite{yeung2022adaptive}, they fine-tuned the previous model on freehand 2D US using cycle consistency to reduce the domain gap. 

Unlike in prior work, in this paper, we propose a method that does not require supervised learning and performs 2D-3D US matching by directly inferring the original freehand 2D US. This allows us to better handle the different image appearances between actual freehand 2D US and resliced 2D US from 3D images.
We utilize a novel self-supervised contrastive learning to learn a generalizable image representation for image retrieval in neck US. 
Our contributions include: 
(1) We present a novel contrastive learning approach that leverages intra-sweep similarity to enhance representation learning for improved image retrieval in ultrasound sweep videos. 
Specifically, our method utilizes the temporal sequence of ultrasound image frames and probe positions to allow the model to capture semantic similarities between frames without the need for manual labeling; 
(2) To the best of our knowledge, this is the first work demonstrating the feasibility of using image retrieval in neck US localization system on real patient data.

\section{Methods}
\begin{figure}
\centering
\includegraphics[width=0.94\textwidth]{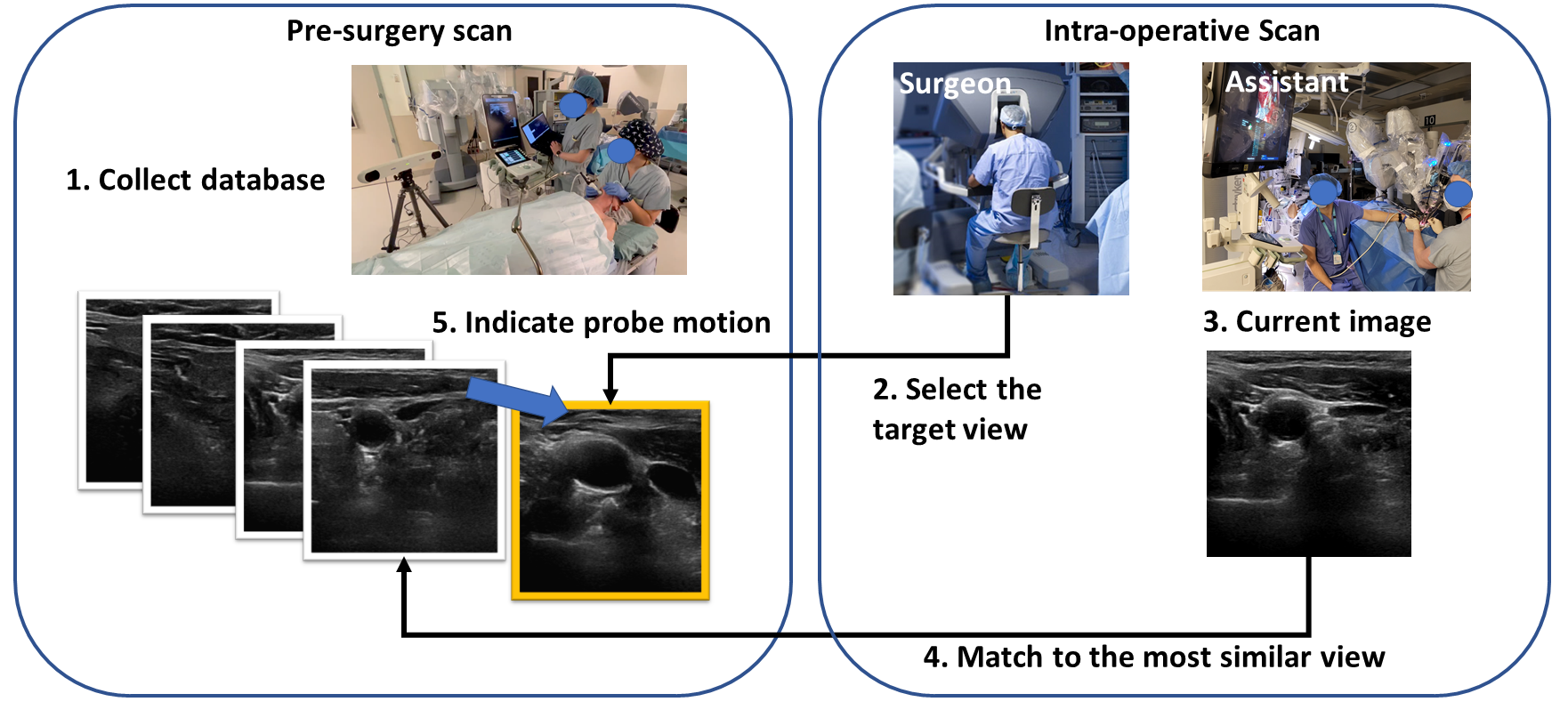}
\caption{The proposed workflow of using image retrieval in US scanning guidance. The surgeon can select the target view, and our method will match the current US image to the most similar view in the database to indicate the current probe location and the desired probe motion. This can provide the US scanning guidance to the surgical assistant, without using an external tracking system during the surgery.}\label{fig:workflow}
\end{figure}

\textbf{Image retrieval for US scanning guidance.} The proposed workflow of using image retrieval for US scanning guidance is shown in Figure~\ref{fig:workflow}. Before the surgery, the surgeon and the surgical assistant perform pre-operative tracked US scanning to collect the image database. 
During the surgery, the surgeon can select the target view from the database, and our model will retrieve the closest frame of the current view from the database. The relative probe location between the retrieved frame and the target frame can be used to indicate the preferred probe motion.
Our main goal is to indicate the large motion required by the operator.

\textbf{Problem description.}
We denote the sweeps as $S^{n}=\{I_0^n,I_1^n,...,I_{T_n}^n\}$, where $I_t^n$ is the frame at time $t$ in sequence $S^n$, and $T_n$ is the length of the $n^{th}$ video sweeps.
The frame encoder, a deep learning model $\mathit{F}$, will project the frame $I_t^n$ into a latent space $\mathbf{z}_t^n=\mathit{F}(I_t^n)$.
Before the surgery, an image sequence $S^{pre}$ is collected on the patient. Given a new US image $I^{intra}$, the feature encoder $\textit{F}$ should retrieve the most similar image $I^{pre}\in S^{pre}$ based on the maximum similarity in the latent space in Eq.~\ref{eq:matching}.
\begin{equation}\label{eq:matching}
    I^{pre} = \arg\max_{I\in S^{pre}} <\mathit{F}(I^{intra}), \mathit{F}(I)>
\end{equation}
Given unlabelled training data $\{S^1,S^2,...,S^N\}$, our method learns the model $\mathit{F}$ needed to evaluate the similarity between the latent embeddings of frames $I$.

\begin{figure}
    \centering
    \includegraphics[width=1.05\textwidth]{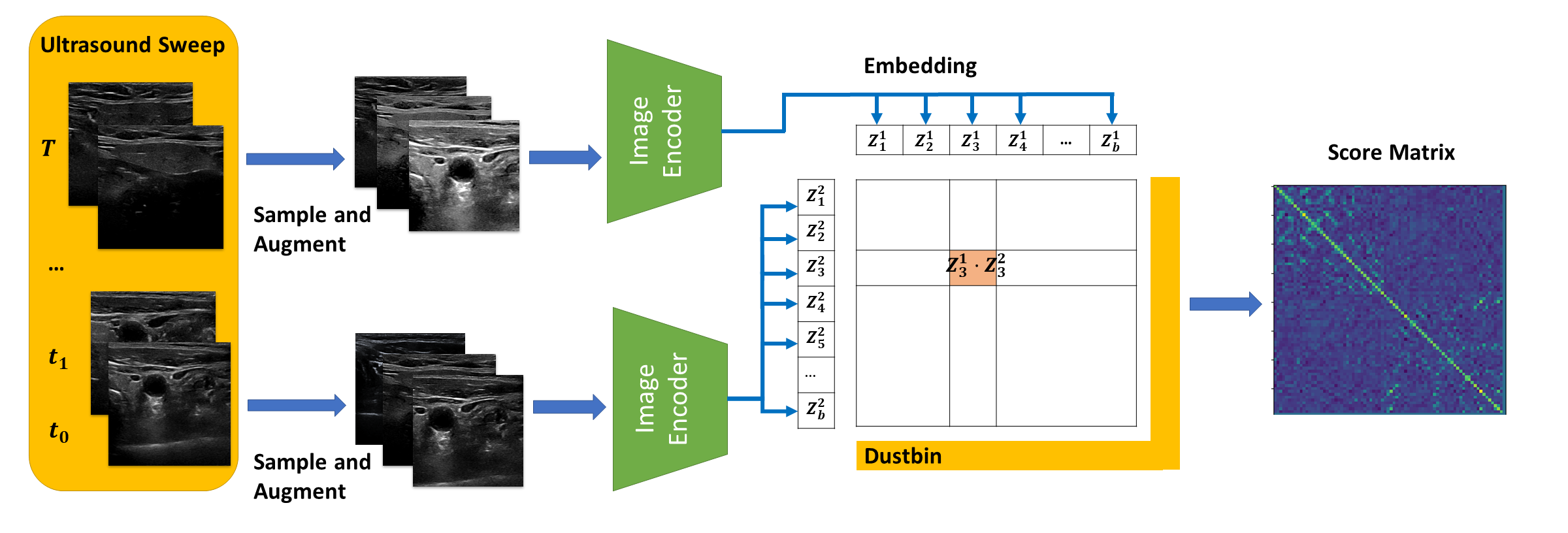}
    \caption{Summary of our intra-sweep training strategy. The image pairs are sampled from one US sweep and augmented to different views, and the image encoder will predict the frame embedding. The dot product of the embedding is used to evaluate the embedding similarity. A dustbin threshold is concatenated to the dot similarity to generate the final score matrix.}\label{fig:model}
\end{figure}

\textbf{Contrastive learning.}
Contrastive learning is a data-efficient method to train a feature encoder. The high-level goal of contrastive learning is to maximize the similarity between the embeddings of positive (or similar) pairs while minimizing the similarity between the embeddings of negative (or dissimilar) pairs.
Our frame-wise representation learning method is summarized in Figure~\ref{fig:model}.
It is inspired by the general contrastive learning framework CLIP~\cite{radford2021learning}. 
CLIP treats representation learning as a view retrieval problem by training the encoders to predict the correct pairings of a batch of training examples, and was originally applied in image-text alignment. For image-image alignment, we use the optimization goal in SimCLRv2~\cite{chen2020simple}. The feature encoder needs to learn a consistent representation of the same image under transformations; therefore, we use a combination of data augmentation on each example twice to create two sets of corresponding views, as illustrated in Fig. \ref{fig:model}.

The main difference between our method and CLIP and SimCLR is that we consider the \textit{intra-sweep semantic similarity} when we draw the samples from the training data. 
CLIP and SimCLR were proposed for large-scale RGB image datasets instead of videos. In their methods, different images are treated as negative pairs and the semantic similarity between images is not considered.
However, this naive method is not optimal for US because US images can capture similar views.
Another naive method is to treat the frames from the same video as positive pairs, and the frames from different videos as negative pairs.
However, operators can revisit the same anatomy in different scans or cover different anatomies in the same sweep. Thus this method fails to capture the semantic similarities and differences between scans.

We propose a new method to consider the \textit{semantic similarity within US sweeps} to guide contrastive learning by utilizing the probe location during the scan.
The probe location is only required in training and is not required during test time.
In training, we have probe locations recorded by an external optical tracker.
The hypothesis is that the semantic meaning of US images, i.e., the anatomy structures inside the images, can be embedded as a function of the US probe location. 
To simplify the problem, we did not consider the probe rotation in this study. In general, the surgeon and the assistant will keep the probe normal to the skin surface.
Therefore we only consider the probe translation in this study and will add probe orientation in future work.
Instead of sampling images from different sweeps in the training stage, we sample the images from the same sweep. 
The image pairs are positive if the probe location distance is smaller than a pre-defined threshold, otherwise, they will be negative. 
If there are multiple positive pairs, the image pair with the smallest distance is positive, and the rest will be negative.
Another difference between our model and CLIP/SimCLR is that we add a mechanism to reject the retrieval when the model is not confident about the retrieved results, which is important in clinical applications. 
We add a learnable threshold to reject the retrieval if the embedding similarity is lower than the threshold. This is referred to as a dustbin. 

To learn the intra-sweep frame representation, in each batch, we sample two \textit{different} batches of frames from the same sweep.
This is to enable the training of the dustbin, so that the model learns a threshold value of the embedding similarity to reject uncertain retrieval. 
We first sample $b$ samples $Batch_1=\{I_{t_1}^n, ..., I_{t_b}^n\}$ from sweep $S^n$, and choose 75\% of the samples in $Batch_1$ for $Batch_2$. 
We then sample another 25\% batch size number of samples not in $Batch_1$ to add them into $Batch_2$. 
We then apply random image augmentation on  $Batch_1$ and  $Batch_2$. 
The feature encoder $\mathit{F}$ encodes the frames to generate embeddings $z_{batch}^1$ and $z_{batch}^2$. 
We calculate the dot product between the feature embeddings to generate the matching score matrix. 
We choose not to normalize the embedding before calculating the similarity as suggested in SuperGlue~\cite{sarlin2020superglue}, because the feature magnitude may encode the confidence of the features, and in our experiments, not normalizing the embedding leads to better convergence.
We then concatenate the dustbin threshold to the score matrix to generate the final score matrix $M$.
We use the symmetric cross-entropy loss in CLIP~\cite{radford2021learning}, as described in Eq.~\ref{eq:clip_loss}. $CE$ is the abbreviation of cross-entropy loss. $\tau$ is a temperature parameter to soften the cross-entropy loss. The $gt_{1to2}$ is the label of positive/negative pairs from $batch_1$ to $batch_2$, and the $gt_{2to1}$ is the label of positive/negative pairs from $batch_2$ to $batch_1$.
We also add a triplet loss in Eq.~\ref{eq:triplet_loss} to further pull and push the representation based on the probe location distance $d$.

\begin{equation}\label{eq:clip_loss}
L_{SCE} = [\sum_i^{B_1} CE(M(\dots,i)\times e^\tau, gt_{1to2}) +  \sum_j^{B_2}CE(M(j,\dots)\times e^\tau, gt_{2to1})]/2
\end{equation}

\begin{equation}\label{eq:triplet_loss}
    L_{triplet} = \sum_i^{B_1} \sum_j^{B_2} d_{ij}\times M_{ij} - ( 1 - d_{ij}) \times M_{ji}
\end{equation}

 \section{Experiments}
\textbf{Dataset:}
We used a private dataset containing 2D US sweeps collected from 19 patients who underwent TORS from January 2022 to October 2023 at the Vancouver General Hospital (Vancouver, BC, Canada). This study received ethics approval from the UBC Clinical Research Ethics Board (H19-04025).
A BK3500 and a 14L3 linear 2D transducer (BK Medical, Burlington, MA) were used in the operation room for US imaging and a Polaris Spectra (Northern Digital, ON, Canada) was used to track the US transducer. 
PLUS~\cite{Lasso2014a} was used to record the US videos.
The image depth is 4 or 5~cm at 9 MHz, with a frame rate of $5.76\pm0.89$ fps.
For each patient, the US scan included the neck on the cancerous side, before and after the tongue retraction. 
The summary of data is shown in Table~\ref{tab:dataset}.

\begin{table}[]
\caption{Dataset summary.}~\label{tab:dataset}
\begin{tabular}{cccc}
\toprule
Dataset    & Number of Patients & Number of Videos & Number of Frames \\
\midrule
Training   & 9                  & 42               & 5353  \\
Validation & 4                  & 24               & 2638  \\
Testing    & 5                  & 27               & 3475  \\
\bottomrule
\end{tabular}
\end{table}

 \textbf{Implementation details:} The feature encoder includes a CNN backbone and a MLP model. The CNN backbone is a ResNet18 without the last pooling and fully connected layers  The MLP model includes 4 linear layers following  ReLU activation and 1DBatchNorm. The number of neurons at each layer is 512. The models were trained on a 16GB Nvidia Tesla V100 and implemented in Python 3.8, PyTorch-2.1.0, and CUDA-11.8. The weight of the encoder was initialized randomly. The Adam optimizer was used with a learning rate of 1e-3, and a StepLR scheduler was used with step size 100 and $\gamma=0.95$. The maximum epoch is 300, and the models with the lowest loss on the validation set were selected. The sampling size for each batch is 30. The temperature parameter $\tau$ is set as 0.1. The distance threshold for a positive pair is $1cm$, and the dustbin value was initialized as 0. The data augmentation includes random affine transformation, resize crop, and color jittering.

\section{Results and Discussion}

\begin{figure}
    \centering
    \includegraphics[width=\textwidth]{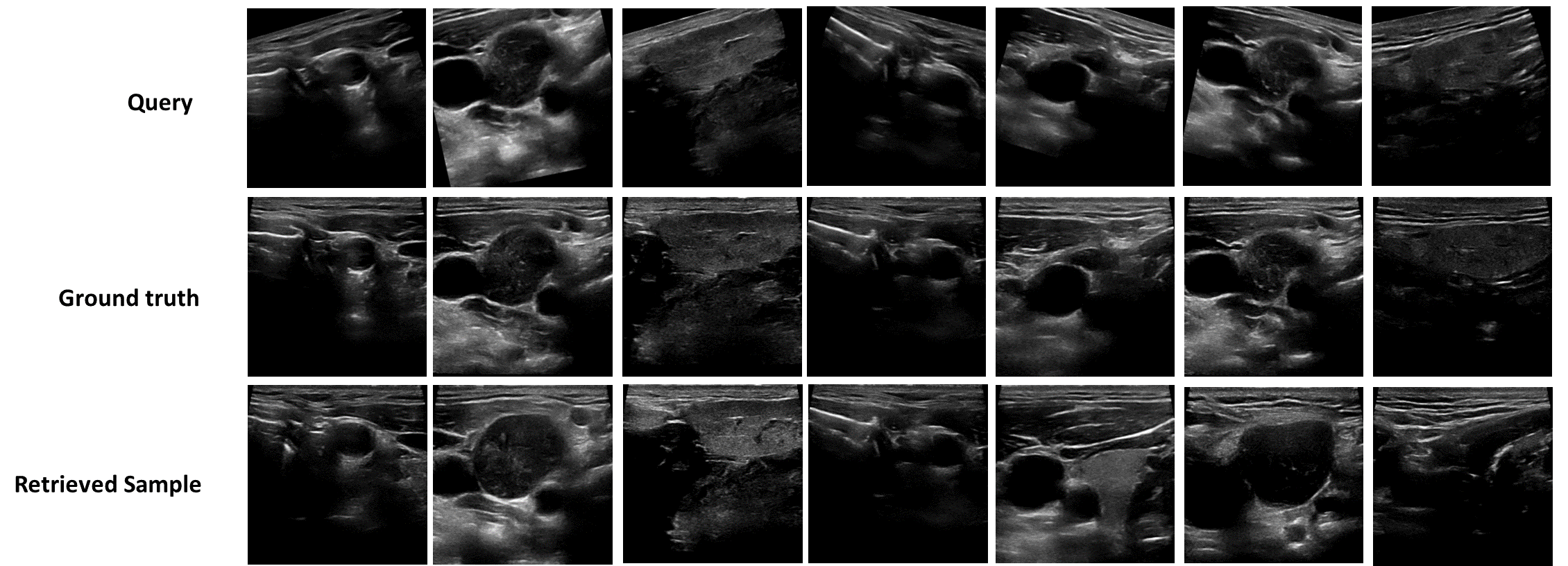}
    \caption{Example of the retrieved frames using our proposed method. The first four columns show correct retrievals and the last three columns show inaccurate matches.}\label{fig:simulate_example}
\end{figure}

\textbf{Simulation study.}
For each testing sweep, we randomly sampled 50 frames from the sweep and transformed them using data augmentation different from training for testing.
To simulate the change of probe orientation, we concatenated the sampled frames with 30 frames temporally before and after it to form a ``mini-volume'' and performed 3D affine augmentation, instead of a pure 2D augmentation. The transform can introduce out-of-plane views.
We compare our method with four baselines. 
The first is to retrieve the image with the highest normalized cross-correlation (NCC).
The second is inter-sweep contrastive learning (inter-sweep CL) using symmetric cross-entropy loss. 
The training dataset consists of image frames obtained from the training US sweeps, and the batches are randomly sampled from this dataset.
Augmented views create a positive pair when they originate from the same image.
This is to compare our intra-sweep strategy with regular inter-sweep sampling.
The third baseline is Intra-Video Positive Pairs (IVPP)~\cite{vanberlo2024intra}, a state-of-the-art intra-video representation learning method. 
IVPP adds the sample weight based on the temporal difference in the video to penalize mismatched positive pairs rather than negative pairs. 
The sampling weight is given by $w=(\delta_t-|t_2-t_1|)/(\delta_t+1)$, where $\delta_t$ is the maximum separation, in the number of frames, in a positive pair.
We set $\delta_t=8$.
The fourth baseline method is distance-IVPP, which is modified from IVPP to use the difference between the probe locations of the two samples as the sample weight.
We set the sampling weight as $w=(\delta_{probe}-||p_2-p_1||_2)/(\delta_{probe}+1)$, where $p_1,p_2$ represent the probe location at $t_1$ and $t_2$. We set $\delta_{probe}=10mm$.
We didn't compare our method with SimCLRv2 because its loss function can not be applied to the dustbin.
We report the retrieval success rate. Image retrieval is successful if the probe location difference between the ground truth and the retrieved image is smaller than 15mm, as defined in~\cite{zhao2021visual}.
We also report the average L2 distance of the probe location difference for the retrieval attempts that are not rejected. 
The quantitative results are shown in Table~\ref{tab:baseline}. 
Our method outperforms the baseline methods, achieving the highest success rate in image retrieval and the lowest probe localization error.
Compared to inter-sweep CL, our approach effectively leverages the probe location's weak signals to enhance the learned feature encoder.
Notably, the inter-sweep CL outperforms intra-sweep sample weighting methods IVPP and distance-IVPP using the intra-sweep sampling strategy. 
We hypothesize that the symmetric cross-entropy loss enforces one-to-one matches to recognize whether the transformed images originate from the same sample.
Therefore, it penalizes the image pairs that actually appear similar but are negative matches. 
Though IVPP and distance-IVPP reduce the penalty on negative pairs, it may not be efficient enough, potentially leading the model to converge to a sub-optimal solution in our data.
In contrast, the inter-sweep CL can draw samples from different sweeps, reducing the likelihood of encountering multiple similar samples within a batch. Thus, the one-to-one matching still enables the model to learn a well-representative embedding.
The results highlight the critical role of the optimization goal in contrastive representation learning.
We provide a visualization of the retrieval results of our method in Figure~\ref{fig:simulate_example}. We can see that the successfully retrieved images localize the correct anatomy. We also demonstrate inaccurate retrieval results, where the models are confused by similar structures in the image. 

\begin{table}[]
\centering
\caption{Quantitative results based on the simulation study for baseline comparisons.}\label{tab:baseline}
\begin{tabular}{cccc}
\toprule
Methods                       & Retrieval success rate   & Probe distance (mm)        & Rejected retrieval \\
\midrule
NCC                           & 68.81\%           & 13.66  $\pm$  17.85        & 0.00\%          \\
Inter-sweep CL           & 90.67\%           & 6.14   $\pm$  15.24        & 0.00\%          \\
IVPP~\cite{vanberlo2024intra} & 83.26\%           & 8.50   $\pm$  17.11        & 3.19\%          \\
Distance-IVPP                 & 83.66\%           & 8.19   $\pm$  14.86        & 1.85\%          \\
Ours                          & \textbf{92.30\%}  & \textbf{5.02 $\pm$  10.89} & 0.00\%          \\
\bottomrule
\end{tabular}
\end{table}

We conduct an ablation study on our proposed method to evaluate the necessity of each design stage. The results in Table~\ref{tab:ablation} show that using the probe location to define positive/negative pairs improves the symmetric cross-entropy loss, meaning that the probe location difference can be a good indicator of frame-level similarity. The triplet loss improves the symmetric cross-entropy loss because it can pull the representation of positive pairs together and push the representation of negative pairs away. 

\begin{table}[]
\caption{Ablation study of the main components in our proposed loss. SCE: Symmetric cross-entropy loss,  P1: With probe location as positive/negative indicator, P2: triplet loss.}\label{tab:ablation}
\begin{tabular}{cccc}
\toprule
              & Retrieval success rate    & Probe distance (mm)         & Rejected retrieval \\
\midrule
SCE loss      & 86.03\%            & 6.52  $\pm$   10.54         & 5.11\%          \\
+P1           & 89.26\%            & 5.77  $\pm$   13.63         & 0.00\%          \\
+P2           & 87.46\%            & 6.58  $\pm$   13.70         & 2.52\%          \\
+P1+P2 (Ours) & \textbf{92.30\%  } & \textbf{5.02 $\pm$ 10.89 }  & 0.00\%          \\
\bottomrule
\end{tabular}
\end{table}

\begin{figure}
    \centering
    \includegraphics[width=\textwidth]{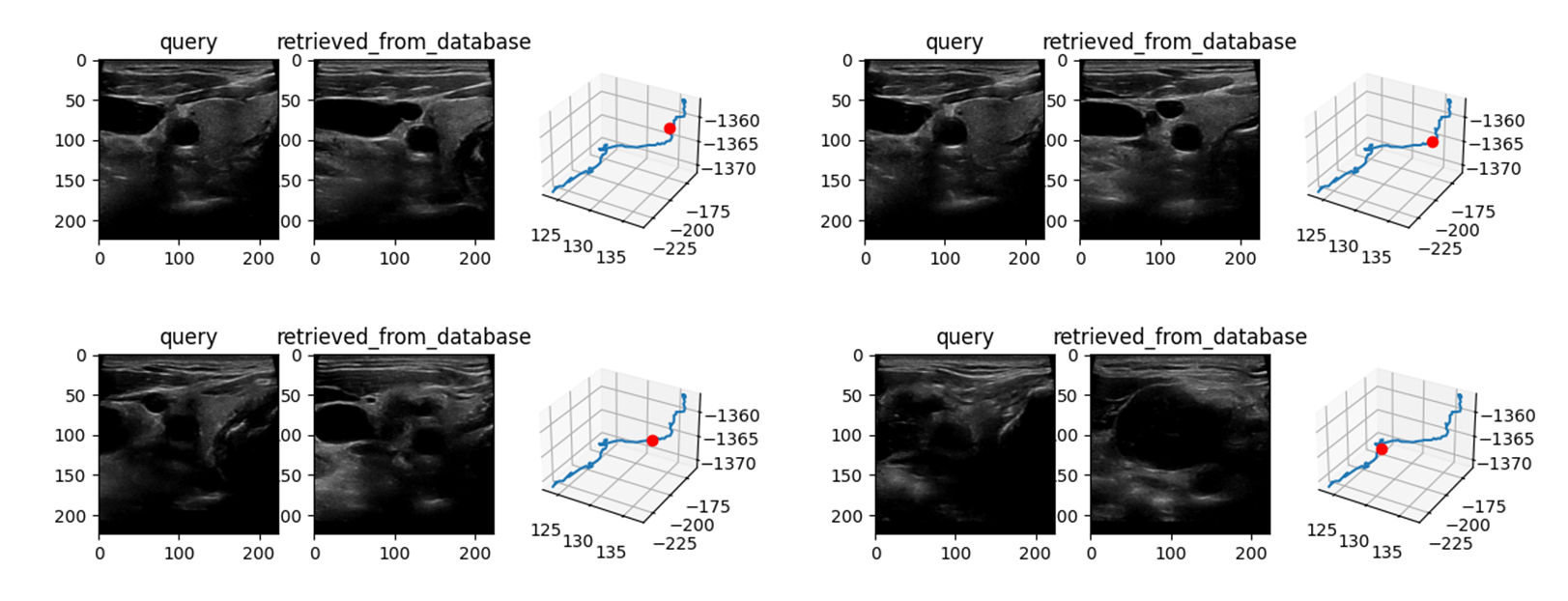}
    \caption{Queries are samples from the post-retraction US, and the database is the pre-retraction US. The blue trajectory is the scanning trajectory in the pre-retraction scan, and the red dot is the localized probe location based on the image retrieval.}\label{fig:cross_sweep_localization}
\end{figure}

\begin{figure}
    \centering
    \includegraphics[width=0.95\textwidth]{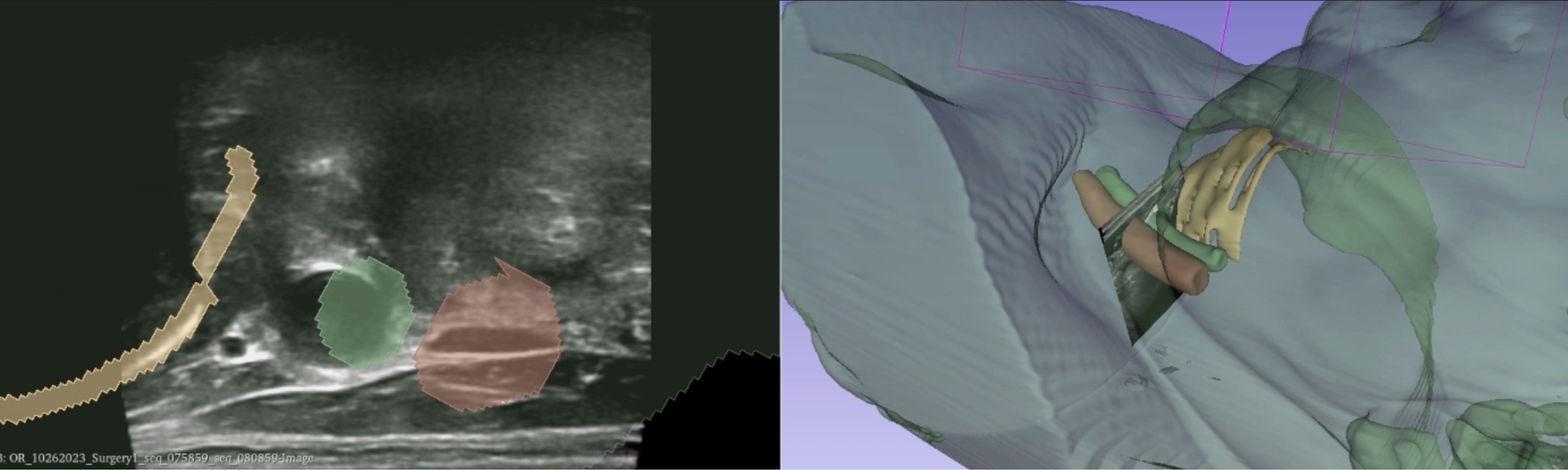}
    \caption{Illustration of the TORS guidance application. The US image is after tongue retraction, and the 3D patient model is extracted from pre-operative CT and aligned with the pre-operative 3D US. The segmentation (yellow: larynx cartilage, green: carotid, red: jugular vein) is roughly aligned with the 2D US. }\label{fig:guidance_demo}
\end{figure}

\textbf{Patient study.}
We provide a proof-of-concept demonstration of intra-operative US probe localization using image retrieval as shown in Figure~\ref{fig:cross_sweep_localization}. 
Tongue retraction is a procedure to pull out the patient's tongue to enhance the exposure of the oropharynx but will cause large tissue deformation. 
The database is US sweep before the tongue retraction, and the query images are from the US sweep collected after the tongue retraction.
The results in Figure~\ref{fig:cross_sweep_localization} show that the retrieved images contain similar anatomy, and we can use the retrieved images to roughly localize the query image. Figure~\ref{fig:guidance_demo} demonstrates our proposed US localization system.

\textbf{Limitations and future work.} While we investigated one dataset and one backbone, the promising results demonstrated the feasibility of our method. 
In the future, we will work on exploring additional feature encoder architectures to enhance performance and generalizability. 
To improve model performance, we will explore adding 3D information into the model, since 2D images do not contain information in the elevational direction.
While we only report the quantitative results in a simulation study - consistent with prior research~\cite{zhao2021visual}, the results inform us of the next steps towards clinical translation.
In this study, we provide only qualitative results on the real-world patient study due to the challenges in directly comparing probe distance before and after tongue retraction. 
However, this study serves as an important proof-of-concept, laying the groundwork for future clinical investigations and the refinement of our method under real-world conditions.

\section{Conclusion}
In this work, we explored the use of image retrieval to guide neck US scanning. We introduced a novel self-supervised contrastive learning strategy that utilizes intra-sweep similarity and probe location information. Our method enhances the performance of the feature encoder and outperforms the state-of-the-art intra-sweep similarity-based representation learning methods. Furthermore, we are the first work demonstrating the feasibility of localizing the US probe in neck US scans by image retrieval and referring to the 2D US only, highlighting the potential of our approach for practical clinical application.

\bmhead{Acknowledgements}
The work is supported by NSERC Discovery Grant and Charles Laszlo Chair in Biomedical Engineering held by Dr. Salcudean, and by VCHRI Innovation and Translational Research Awards and the University of British Columbia Department of Surgery Seed Grant held by Dr. Prisman. 


\section*{Declarations}
\textbf{Competing interests:} The authors have no conflicts of interest.
\textbf{Ethics:} Institutional ethics approval (\#H19-04025) was obtained for this study. Informed consent was obtained from all participants.  The procedures used in this study adhere to the tenets of the Declaration of Helsinki.

\bibliography{sn-bibliography}


\begin{thebibliography}{21}
\ifx \bisbn   \undefined \def \bisbn  #1{ISBN #1}\fi
\ifx \binits  \undefined \def \binits#1{#1}\fi
\ifx \bauthor  \undefined \def \bauthor#1{#1}\fi
\ifx \batitle  \undefined \def \batitle#1{#1}\fi
\ifx \bjtitle  \undefined \def \bjtitle#1{#1}\fi
\ifx \bvolume  \undefined \def \bvolume#1{\textbf{#1}}\fi
\ifx \byear  \undefined \def \byear#1{#1}\fi
\ifx \bissue  \undefined \def \bissue#1{#1}\fi
\ifx \bfpage  \undefined \def \bfpage#1{#1}\fi
\ifx \blpage  \undefined \def \blpage #1{#1}\fi
\ifx \burl  \undefined \def \burl#1{\textsf{#1}}\fi
\ifx \doiurl  \undefined \def \doiurl#1{\url{https://doi.org/#1}}\fi
\ifx \betal  \undefined \def \betal{\textit{et al.}}\fi
\ifx \binstitute  \undefined \def \binstitute#1{#1}\fi
\ifx \binstitutionaled  \undefined \def \binstitutionaled#1{#1}\fi
\ifx \bctitle  \undefined \def \bctitle#1{#1}\fi
\ifx \beditor  \undefined \def \beditor#1{#1}\fi
\ifx \bpublisher  \undefined \def \bpublisher#1{#1}\fi
\ifx \bbtitle  \undefined \def \bbtitle#1{#1}\fi
\ifx \bedition  \undefined \def \bedition#1{#1}\fi
\ifx \bseriesno  \undefined \def \bseriesno#1{#1}\fi
\ifx \blocation  \undefined \def \blocation#1{#1}\fi
\ifx \bsertitle  \undefined \def \bsertitle#1{#1}\fi
\ifx \bsnm \undefined \def \bsnm#1{#1}\fi
\ifx \bsuffix \undefined \def \bsuffix#1{#1}\fi
\ifx \bparticle \undefined \def \bparticle#1{#1}\fi
\ifx \barticle \undefined \def \barticle#1{#1}\fi
\bibcommenthead
\ifx \bconfdate \undefined \def \bconfdate #1{#1}\fi
\ifx \botherref \undefined \def \botherref #1{#1}\fi
\ifx \url \undefined \def \url#1{\textsf{#1}}\fi
\ifx \bchapter \undefined \def \bchapter#1{#1}\fi
\ifx \bbook \undefined \def \bbook#1{#1}\fi
\ifx \bcomment \undefined \def \bcomment#1{#1}\fi
\ifx \oauthor \undefined \def \oauthor#1{#1}\fi
\ifx \citeauthoryear \undefined \def \citeauthoryear#1{#1}\fi
\ifx \endbibitem  \undefined \def \endbibitem {}\fi
\ifx \bconflocation  \undefined \def \bconflocation#1{#1}\fi
\ifx \arxivurl  \undefined \def \arxivurl#1{\textsf{#1}}\fi
\csname PreBibitemsHook\endcsname

\bibitem[\protect\citeauthoryear{Adelstein et~al.}{2017}]{adelstein2017nccn}
\begin{barticle}
\bauthor{\bsnm{Adelstein}, \binits{D.}},
\bauthor{\bsnm{Gillison}, \binits{M.L.}},
\bauthor{\bsnm{Pfister}, \binits{D.G.}},
\bauthor{\bsnm{Spencer}, \binits{S.}},
\bauthor{\bsnm{Adkins}, \binits{D.}},
\bauthor{\bsnm{Brizel}, \binits{D.M.}},
\bauthor{\bsnm{Burtness}, \binits{B.}},
\bauthor{\bsnm{Busse}, \binits{P.M.}},
\bauthor{\bsnm{Caudell}, \binits{J.J.}},
\bauthor{\bsnm{Cmelak}, \binits{A.J.}}, \betal:
\batitle{Nccn guidelines insights: head and neck cancers, version 2.2017}.
\bjtitle{Journal of the National Comprehensive Cancer Network}
\bvolume{15}(\bissue{6}),
\bfpage{761}--\blpage{770}
(\byear{2017})
\end{barticle}
\endbibitem

\bibitem[\protect\citeauthoryear{Quan et~al.}{2021}]{quan2021gastrostomy}
\begin{barticle}
\bauthor{\bsnm{Quan}, \binits{D.L.}},
\bauthor{\bsnm{Sukari}, \binits{A.}},
\bauthor{\bsnm{Nagasaka}, \binits{M.}},
\bauthor{\bsnm{Kim}, \binits{H.}},
\bauthor{\bsnm{Cramer}, \binits{J.D.}}:
\batitle{Gastrostomy tube dependence and patient-reported quality of life outcomes based on type of treatment for human papillomavirus-associated oropharyngeal cancer: systematic review and meta-analysis}.
\bjtitle{Head \& Neck}
\bvolume{43}(\bissue{11}),
\bfpage{3681}--\blpage{3696}
(\byear{2021})
\end{barticle}
\endbibitem

\bibitem[\protect\citeauthoryear{Lechner et~al.}{2022}]{lechner2022hpv}
\begin{barticle}
\bauthor{\bsnm{Lechner}, \binits{M.}},
\bauthor{\bsnm{Liu}, \binits{J.}},
\bauthor{\bsnm{Masterson}, \binits{L.}},
\bauthor{\bsnm{Fenton}, \binits{T.R.}}:
\batitle{Hpv-associated oropharyngeal cancer: epidemiology, molecular biology and clinical management}.
\bjtitle{Nature reviews Clinical oncology}
\bvolume{19}(\bissue{5}),
\bfpage{306}--\blpage{327}
(\byear{2022})
\end{barticle}
\endbibitem

\bibitem[\protect\citeauthoryear{Liu et~al.}{2015}]{liu2015augmented}
\begin{barticle}
\bauthor{\bsnm{Liu}, \binits{W.P.}},
\bauthor{\bsnm{Richmon}, \binits{J.D.}},
\bauthor{\bsnm{Sorger}, \binits{J.M.}},
\bauthor{\bsnm{Azizian}, \binits{M.}},
\bauthor{\bsnm{Taylor}, \binits{R.H.}}:
\batitle{Augmented reality and cone beam ct guidance for transoral robotic surgery}.
\bjtitle{Journal of robotic surgery}
\bvolume{9},
\bfpage{223}--\blpage{233}
(\byear{2015})
\end{barticle}
\endbibitem

\bibitem[\protect\citeauthoryear{Kahng et~al.}{2019}]{kahng2019improving}
\begin{barticle}
\bauthor{\bsnm{Kahng}, \binits{P.W.}},
\bauthor{\bsnm{Wu}, \binits{X.}},
\bauthor{\bsnm{Ramesh}, \binits{N.P.}},
\bauthor{\bsnm{Pastel}, \binits{D.A.}},
\bauthor{\bsnm{Halter}, \binits{R.J.}},
\bauthor{\bsnm{Paydarfar}, \binits{J.A.}}:
\batitle{Improving target localization during trans-oral surgery with use of intraoperative imaging}.
\bjtitle{International Journal of Computer Assisted Radiology and Surgery}
\bvolume{14},
\bfpage{885}--\blpage{893}
(\byear{2019})
\end{barticle}
\endbibitem

\bibitem[\protect\citeauthoryear{Chen et~al.}{2023}]{chen2023towards}
\begin{barticle}
\bauthor{\bsnm{Chen}, \binits{W.}},
\bauthor{\bsnm{Kalia}, \binits{M.}},
\bauthor{\bsnm{Zeng}, \binits{Q.}},
\bauthor{\bsnm{Pang}, \binits{E.H.}},
\bauthor{\bsnm{Bagherinasab}, \binits{R.}},
\bauthor{\bsnm{Milner}, \binits{T.D.}},
\bauthor{\bsnm{Sabiq}, \binits{F.}},
\bauthor{\bsnm{Prisman}, \binits{E.}},
\bauthor{\bsnm{Salcudean}, \binits{S.E.}}:
\batitle{Towards transcervical ultrasound image guidance for transoral robotic surgery}.
\bjtitle{International Journal of Computer Assisted Radiology and Surgery}
\bvolume{18}(\bissue{6}),
\bfpage{1061}--\blpage{1068}
(\byear{2023})
\end{barticle}
\endbibitem

\bibitem[\protect\citeauthoryear{Green et~al.}{2020}]{green2020integrated}
\begin{barticle}
\bauthor{\bsnm{Green}, \binits{E.D.}},
\bauthor{\bsnm{Paleri}, \binits{V.}},
\bauthor{\bsnm{Hardman}, \binits{J.C.}},
\bauthor{\bsnm{Kerawala}, \binits{C.}},
\bauthor{\bsnm{Riva}, \binits{F.M.}},
\bauthor{\bsnm{Jaly}, \binits{A.A.}},
\bauthor{\bsnm{Ap~Dafydd}, \binits{D.}}:
\batitle{Integrated surgery and radiology: trans-oral robotic surgery guided by real-time radiologist-operated intraoral ultrasound}.
\bjtitle{Oral and Maxillofacial Surgery}
\bvolume{24},
\bfpage{477}--\blpage{483}
(\byear{2020})
\end{barticle}
\endbibitem

\bibitem[\protect\citeauthoryear{Chang et~al.}{2021}]{chang2021real}
\begin{barticle}
\bauthor{\bsnm{Chang}, \binits{C.-C.}},
\bauthor{\bsnm{Wu}, \binits{J.-L.}},
\bauthor{\bsnm{Hsiao}, \binits{J.-R.}},
\bauthor{\bsnm{Lin}, \binits{C.-Y.}}:
\batitle{Real-time, intraoperative, ultrasound-assisted transoral robotic surgery for obstructive sleep apnea}.
\bjtitle{The Laryngoscope}
\bvolume{131}(\bissue{4}),
\bfpage{1383}--\blpage{1390}
(\byear{2021})
\end{barticle}
\endbibitem

\bibitem[\protect\citeauthoryear{Moore et~al.}{2024}]{moore2024enabling}
\begin{botherref}
\oauthor{\bsnm{Moore}, \binits{R.}},
\oauthor{\bsnm{Yeung}, \binits{R.}},
\oauthor{\bsnm{Chen}, \binits{W.}},
\oauthor{\bsnm{Zeng}, \binits{Q.}},
\oauthor{\bsnm{Prisman}, \binits{E.}},
\oauthor{\bsnm{Salcudean}, \binits{S.}}:
Enabling extracorporeal ultrasound imaging with the da vinci robot for transoral robotic surgery: a feasibility study.
International Journal of Computer Assisted Radiology and Surgery,
1--8
(2024)
\end{botherref}
\endbibitem

\bibitem[\protect\citeauthoryear{Grimwood et~al.}{2020}]{grimwood2020assisted}
\begin{bchapter}
\bauthor{\bsnm{Grimwood}, \binits{A.}},
\bauthor{\bsnm{McNair}, \binits{H.}},
\bauthor{\bsnm{Hu}, \binits{Y.}},
\bauthor{\bsnm{Bonmati}, \binits{E.}},
\bauthor{\bsnm{Barratt}, \binits{D.}},
\bauthor{\bsnm{Harris}, \binits{E.J.}}:
\bctitle{Assisted probe positioning for ultrasound guided radiotherapy using image sequence classification}.
In: \bbtitle{International Conference on Medical Image Computing and Computer-Assisted Intervention},
pp. \bfpage{544}--\blpage{552}
(\byear{2020}).
\bcomment{Springer}
\end{bchapter}
\endbibitem

\bibitem[\protect\citeauthoryear{Droste et~al.}{2020}]{droste2020automatic}
\begin{bchapter}
\bauthor{\bsnm{Droste}, \binits{R.}},
\bauthor{\bsnm{Drukker}, \binits{L.}},
\bauthor{\bsnm{Papageorghiou}, \binits{A.T.}},
\bauthor{\bsnm{Noble}, \binits{J.A.}}:
\bctitle{Automatic probe movement guidance for freehand obstetric ultrasound}.
In: \bbtitle{Medical Image Computing and Computer Assisted Intervention--MICCAI 2020: 23rd International Conference, Lima, Peru, October 4--8, 2020, Proceedings, Part III 23},
pp. \bfpage{583}--\blpage{592}
(\byear{2020}).
\bcomment{Springer}
\end{bchapter}
\endbibitem

\bibitem[\protect\citeauthoryear{Zhao et~al.}{2022}]{zhao2022uspoint}
\begin{bchapter}
\bauthor{\bsnm{Zhao}, \binits{C.}},
\bauthor{\bsnm{Droste}, \binits{R.}},
\bauthor{\bsnm{Drukker}, \binits{L.}},
\bauthor{\bsnm{Papageorghiou}, \binits{A.T.}},
\bauthor{\bsnm{Noble}, \binits{J.A.}}:
\bctitle{Uspoint: Self-supervised interest point detection and description for ultrasound-probe motion estimation during fine-adjustment standard fetal plane finding}.
In: \bbtitle{International Conference on Medical Image Computing and Computer-Assisted Intervention},
pp. \bfpage{104}--\blpage{114}
(\byear{2022}).
\bcomment{Springer}
\end{bchapter}
\endbibitem

\bibitem[\protect\citeauthoryear{Men et~al.}{2024}]{men2024pose}
\begin{botherref}
\oauthor{\bsnm{Men}, \binits{Q.}},
\oauthor{\bsnm{Guo}, \binits{X.}},
\oauthor{\bsnm{Papageorghiou}, \binits{A.T.}},
\oauthor{\bsnm{Noble}, \binits{J.A.}}:
Pose-guidenet: Automatic scanning guidance for fetal head ultrasound from pose estimation.
arXiv preprint arXiv:2408.09931
(2024)
\end{botherref}
\endbibitem

\bibitem[\protect\citeauthoryear{Zhao et~al.}{2021}]{zhao2021visual}
\begin{bchapter}
\bauthor{\bsnm{Zhao}, \binits{C.}},
\bauthor{\bsnm{Droste}, \binits{R.}},
\bauthor{\bsnm{Drukker}, \binits{L.}},
\bauthor{\bsnm{Papageorghiou}, \binits{A.T.}},
\bauthor{\bsnm{Noble}, \binits{J.A.}}:
\bctitle{Visual-assisted probe movement guidance for obstetric ultrasound scanning using landmark retrieval}.
In: \bbtitle{Medical Image Computing and Computer Assisted Intervention--MICCAI 2021: 24th International Conference, Strasbourg, France, September 27--October 1, 2021, Proceedings, Part VIII 24},
pp. \bfpage{670}--\blpage{679}
(\byear{2021}).
\bcomment{Springer}
\end{bchapter}
\endbibitem

\bibitem[\protect\citeauthoryear{Yeung et~al.}{2021}]{yeung2021learning}
\begin{barticle}
\bauthor{\bsnm{Yeung}, \binits{P.-H.}},
\bauthor{\bsnm{Aliasi}, \binits{M.}},
\bauthor{\bsnm{Papageorghiou}, \binits{A.T.}},
\bauthor{\bsnm{Haak}, \binits{M.}},
\bauthor{\bsnm{Xie}, \binits{W.}},
\bauthor{\bsnm{Namburete}, \binits{A.I.}}:
\batitle{Learning to map 2d ultrasound images into 3d space with minimal human annotation}.
\bjtitle{Medical Image Analysis}
\bvolume{70},
\bfpage{101998}
(\byear{2021})
\end{barticle}
\endbibitem

\bibitem[\protect\citeauthoryear{Yeung et~al.}{2022}]{yeung2022adaptive}
\begin{bchapter}
\bauthor{\bsnm{Yeung}, \binits{P.-H.}},
\bauthor{\bsnm{Aliasi}, \binits{M.}},
\bauthor{\bsnm{Haak}, \binits{M.}},
\bauthor{\bsnm{Consortium}, \binits{I.-s.}},
\bauthor{\bsnm{Xie}, \binits{W.}},
\bauthor{\bsnm{Namburete}, \binits{A.I.}}:
\bctitle{Adaptive 3d localization of 2d freehand ultrasound brain images}.
In: \bbtitle{International Conference on Medical Image Computing and Computer-Assisted Intervention},
pp. \bfpage{207}--\blpage{217}
(\byear{2022}).
\bcomment{Springer}
\end{bchapter}
\endbibitem

\bibitem[\protect\citeauthoryear{Radford et~al.}{2021}]{radford2021learning}
\begin{bchapter}
\bauthor{\bsnm{Radford}, \binits{A.}},
\bauthor{\bsnm{Kim}, \binits{J.W.}},
\bauthor{\bsnm{Hallacy}, \binits{C.}},
\bauthor{\bsnm{Ramesh}, \binits{A.}},
\bauthor{\bsnm{Goh}, \binits{G.}},
\bauthor{\bsnm{Agarwal}, \binits{S.}},
\bauthor{\bsnm{Sastry}, \binits{G.}},
\bauthor{\bsnm{Askell}, \binits{A.}},
\bauthor{\bsnm{Mishkin}, \binits{P.}},
\bauthor{\bsnm{Clark}, \binits{J.}}, \betal:
\bctitle{Learning transferable visual models from natural language supervision}.
In: \bbtitle{International Conference on Machine Learning},
pp. \bfpage{8748}--\blpage{8763}
(\byear{2021}).
\bcomment{PMLR}
\end{bchapter}
\endbibitem

\bibitem[\protect\citeauthoryear{Chen et~al.}{2020}]{chen2020simple}
\begin{bchapter}
\bauthor{\bsnm{Chen}, \binits{T.}},
\bauthor{\bsnm{Kornblith}, \binits{S.}},
\bauthor{\bsnm{Norouzi}, \binits{M.}},
\bauthor{\bsnm{Hinton}, \binits{G.}}:
\bctitle{A simple framework for contrastive learning of visual representations}.
In: \bbtitle{International Conference on Machine Learning},
pp. \bfpage{1597}--\blpage{1607}
(\byear{2020}).
\bcomment{PMLR}
\end{bchapter}
\endbibitem

\bibitem[\protect\citeauthoryear{Sarlin et~al.}{2020}]{sarlin2020superglue}
\begin{bchapter}
\bauthor{\bsnm{Sarlin}, \binits{P.-E.}},
\bauthor{\bsnm{DeTone}, \binits{D.}},
\bauthor{\bsnm{Malisiewicz}, \binits{T.}},
\bauthor{\bsnm{Rabinovich}, \binits{A.}}:
\bctitle{Superglue: Learning feature matching with graph neural networks}.
In: \bbtitle{Proceedings of the IEEE/CVF Conference on Computer Vision and Pattern Recognition},
pp. \bfpage{4938}--\blpage{4947}
(\byear{2020})
\end{bchapter}
\endbibitem

\bibitem[\protect\citeauthoryear{Lasso et~al.}{2014}]{Lasso2014a}
\begin{barticle}
\bauthor{\bsnm{Lasso}, \binits{A.}},
\bauthor{\bsnm{Heffter}, \binits{T.}},
\bauthor{\bsnm{Rankin}, \binits{A.}},
\bauthor{\bsnm{Pinter}, \binits{C.}},
\bauthor{\bsnm{Ungi}, \binits{T.}},
\bauthor{\bsnm{Fichtinger}, \binits{G.}}:
\batitle{{PLUS}: Open-source toolkit for ultrasound-guided intervention systems}.
\bjtitle{IEEE Transactions on Biomedical Engineering}
\bvolume{61},
\bfpage{2527}--\blpage{2537}
(\byear{2014})
\end{barticle}
\endbibitem

\bibitem[\protect\citeauthoryear{VanBerlo et~al.}{2024}]{vanberlo2024intra}
\begin{botherref}
\oauthor{\bsnm{VanBerlo}, \binits{B.}},
\oauthor{\bsnm{Wong}, \binits{A.}},
\oauthor{\bsnm{Hoey}, \binits{J.}},
\oauthor{\bsnm{Arntfield}, \binits{R.}}:
Intra-video positive pairs in self-supervised learning for ultrasound.
Frontiers in Imaging
\textbf{3}
(2024)
\doiurl{10.3389/fimag.2024.1416114}
\end{botherref}
\endbibitem

\end{thebibliography}

\end{document}